\documentclass{article}
\usepackage{amsmath,graphicx,hyperref}
\usepackage[preprint]{spconf}
\toappear{
\hfill
\begin{minipage}{0.48\textwidth} 
\vspace{-0cm}
\tiny
\copyright\ 2026 IEEE. Personal use of this material is permitted. Permission from IEEE must be obtained for all other uses, in any current or future media, including reprinting/republishing this material for advertising or promotional purposes, creating new collective works, for resale or redistribution to servers or lists, or reuse of any copyrighted component of this work in other works.
\end{minipage}
}

\usepackage{colortbl}
\definecolor{all_range_color}{rgb}{0.17254902, 0.62745098, 0.17254902}
\definecolor{concrete_range_color}{rgb}{1. , 0.49803922, 0.05490196}
\definecolor{abstract_range_color}{rgb}{0.12156863, 0.46666667, 0.70588235}
\usepackage{booktabs}

\title{The impact of abstract and object tags on image privacy classification}

\name{Darya Baranouskaya$^{1, 2}$, Andrea Cavallaro$^{1,2}$}
\address{$^{1}$Idiap Research Institute, Martigny, Switzerland \\ $^{2}$EPFL, Lausanne, Switzerland}

\begin{document}
\ninept 

\maketitle

\begin{abstract}
Object tags denote concrete entities and are central to many computer vision tasks, whereas abstract tags capture higher-level information, which is relevant for tasks that require a contextual, potentially subjective scene understanding. Object and abstract tags extracted from images also facilitate interpretability.  In this paper, we explore which type of tags is more suitable for the context-dependent and inherently subjective task of image privacy. While object tags are generally used for privacy classification, we show that abstract tags are more effective when the tag budget is limited. Conversely, when a larger number of tags per image is available, object-related information is as useful. We believe that these findings will guide future research in developing more accurate image privacy classifiers, informed by the role of tag types and quantity.
\end{abstract}


\section{Introduction}
\label{sec:intro}
Textual information extracted from images by deep learning models (the so-called {\em deep tags}) is commonly used as a natural language description of image content to create interpretable image classifiers~\cite{Ayci_2023_PEAK, 8features, baia2024image} or improve the performance of vision classifiers~\cite{Zhao_2022_privacyalert, Zhao_2023_deep_gated, liu2023modality_coupling, huang2024tag2text, huang2022idea, Stoidis_2022, xompero2025learningprivacyvisualentities, kqiku2024sensitivalert}.
{\em Object} (i.e.~concrete) tags refer to specific physical entities (e.g.~a passport), whereas {\em abstract} tags define for example actions (e.g.~running), qualities (e.g.~old), and higher-level notions (e.g.~spirituality, responsibility, justice)~\cite{brysbaert2014concreteness}.
For object-guided vision tasks, the types of textual information that effectively capture visual concepts are relatively well understood~\cite{huang2024tag2text, yang2023language_model_guided_bottlenecks_for_interpretable_image_classification}. In contrast, limited research has examined which tag types are most suitable for addressing ambiguous and subjective tasks, such as privacy classification~\cite{Li_2020_taxonomy}.

Most of the image privacy classifiers rely on concrete tags such as ImageNet~\cite{ImageNet} class names or scene information~\cite{Zhao_2022_privacyalert, liu2023modality_coupling, Stoidis_2022, xompero2025learningprivacyvisualentities, Tonge_Caragea_Squicciarini_2018}. However, since humans rely on both concrete and abstract concepts when making decisions~\cite{grossmann2024wise_mind_balances_abstarct_and_concrete} and image privacy classification is a subjective task~\cite{Hrynenko_POPETS_2025} that requires complex human-like understanding of the visual content~\cite{habib2019impact_of_context_on_snapchat_public_sharing}, utilising only object information for image description might be limiting. Previous research~\cite{8features} showed that the image classifier built only on eight privacy-specific concepts, which combined abstract information about spoofed, medical, violent, adult or racy content present in the image, together with information about objects (people), can achieve comparable performance to larger and more complex approaches. 

Previous works have demonstrated that people associate diverse words with both abstract and concrete concepts when annotating images~\cite{tater2024evaluating_semantic_relations_in_predicting_textual_labels}. However, very little research has examined how the abstractness of tags influences the performance of classifiers, specifically for image privacy. The most relevant work to ours~\cite{galfre2019exploring_abstract_concepts} showed that the image privacy classifier achieves the best performance with a combination of abstract and concrete tags. 
Yet, this work utilised tags that users assigned to images when posting them on social media. However, user tags are rarely available~\cite{Tonge_Caragea_2020_image_priv_pred_using_deep_nn}, which restricted the previous discussion of abstractness~\cite{galfre2019exploring_abstract_concepts} to a single dataset with only a small number of tags per image. Moreover, user tags significantly differ from automatically generated deep tags~\cite{Zhao_2022_privacyalert, Tonge_Caragea_2020_image_priv_pred_using_deep_nn}, thereby limiting the scalability and applicability of these findings to future research.

In this paper, we study the impact of the tag abstractness on image privacy classification. We investigate how the type and level of detail of deep tags influence the performance of image classifiers across multiple privacy datasets. We identify a relationship between the subjectivity of the classification task and the informativeness of abstract tags, and provide recommendations on which concept types should be utilised in tag-based image privacy classifiers to improve the performance. We ensure a fair comparison between different tag types (concrete vs. abstract) by constraining both the dictionary size and the number of tags used for each image, and explore datasets with both subjective and object-guided annotations. As a result, we show that abstract tags play a key role in the (\textit{subjective}) task of image privacy classification, particularly when only a small number of tags is used to describe an image. In contrast, with a larger number of tags, comparable performance to the one on abstract tags can be achieved using only concrete tags or a combination of the two types. For tasks with {\em{object-guided}} annotation, concrete tags perform better than abstract tags, however, utilising only abstract information for such tasks does not significantly lower the performance.

\section{Data}

We use three public datasets, each having a different approach towards privacy annotation as summarised in Table~\ref{tab: datasets}.

\noindent {\bf PrivacyAlert} contains 6800 images with binary labels (private or public). These labels were derived by aggregating subjective judgments from three to five annotators, who were instructed to assume that the images were taken by themselves and of people they know, and to decide on image privacy. 

\noindent {\bf VISPR} contains 22112 images annotated with respect to 67 private attributes, where the presence of any attribute signals image privacy. A subset of these attributes concerns concrete entities such as different types of documents and vehicles, which are strongly tied to the presence of specific objects in the image. The majority of VISPR attributes relate to personal descriptions, life and relationships, which require the presence of a person in the image. This skews the distribution of private images towards those depicting people~\cite{samson2024little_data_big_impact}. The VISPR annotation is considerably less subjective than PrivacyAlert and is predominantly object-guided.
\begin{table}[t]
\centering
\caption{Datasets and the information about their annotation.}
\vspace{7pt}
\begin{tabular}{p{1.8cm}p{1.3cm}p{4.2cm}}
\toprule
\textbf{Dataset} & \textbf{Annotation type} & \textbf{Annotation details} \\
\midrule
PrivacyAlert~\cite{Zhao_2022_privacyalert} & Subjective &  A judgment of the privacy of the whole image aggregated across a few annotators. \\
VISPR~\cite{Orekondy_2017} & Object-guided &  The annotated presence of an object-guided attribute in an image makes it private.\\
DIPA2~\cite{xu2024dipa2} & Subjective and~object-guided & The privacy of each object in the image is annotated by four annotators, which are aggregated into binary image labels. \\
\bottomrule
\end{tabular}
\label{tab: datasets}
\end{table}

\noindent {\bf DIPA2} contains 1304 images, for which annotators were presented with a list of detected privacy-related objects and their bounding boxes, and asked to rate privacy-related properties for each object. The main properties include \textit{Privacy Threatening} (PT), which indicates if the annotator thinks that the object presents privacy-threatening content, and \textit{Risk Severity} (RS), which marks the perceived severity of the privacy threat posed by an object\footnote{If \textit{PT}=0, \textit{RS} was not annotated; therefore, in such cases, we assign \textit{RS} a zero value.}. Since the dataset does not supply binary privacy labels at the image level, we constructed them as follows: 1) for each image, we took the highest \textit{PT} and \textit{RS} scores assigned by each annotator across all objects in the image, and averaged these values across all four annotators for \textit{PT} and \textit{RS} independently. This procedure resulted in each image receiving two aggregated annotations: \textit{PT} $\in [0, 1]$ and \textit{RS} $\in [0, 7]$.  2) To favour protection, we assigned a final binary public label to an image only if the average \textit{PT} and \textit{RS} values were not higher than the midpoint thresholds of 0.5 and 3.5, respectively. All the other images were labelled as private. As a result, the binary labels received for DIPA2 images are object-guided, as annotators judged privacy risks independently for each object, and the annotations of the most privacy-threatening objects were then propagated into image-level labels. However, unlike VISPR, where the mere presence of an attribute determines privacy, in DIPA2, the privacy annotation of each object is done independently and judged by each annotator.


\section{Tag concreteness}
\label{sec: tags and concreteness}

We introduce the tags chosen for the study and describe how the degree of concreteness (which we treat as the opposite end of the same scale as abstractness) is defined for each tag.

\noindent{\textbf{Tag extraction.}}
To obtain a wide range of descriptive tags for each image, we employ the commercial classifier ClarifAI\footnote{https://clarifai.com/clarifai/main/models/general-image-recognition}. This system has been utilised in the newest interpretable, tag-based privacy classifier, achieving high performance~\cite{Ayci_2023_PEAK}. ClarifAI generates a wide range of tags providing varied information about the image content that covers, for example, objects (e.g. ball, man, table, skin), actions (e.g. laughing, dancing), properties of the objects (e.g. pretty, dark), emotions, and other abstract concepts (e.g. love, religion, wedding).
ClarifAI outputs up to 200 tags for each image selected from a fixed dictionary of $N=6568$ words, $T=\{t_{1}, ...,  t_{N}\}$. Each tag $t_j \in T$  is associated with a probability $p(x_i, t_j)$ of being present in image $x_i$ if outputted by ClarifAI or a zero probability $p(x_i, t_j)=0$ otherwise. 
Hence, an image $x_i$ can be represented as a probability vector:
\begin{equation}
    p(x_i, T) = (p(x_i, t_{1}), ...,  p(x_i, t_{N})).
\end{equation}

\noindent{\textbf{Tag concreteness.}}
To measure the concreteness of tags, we rely on the ranking provided by Brysbaert et al.~\cite{brysbaert2014concreteness} for a set $C$ of 40000 tags\footnote{When defining $T$, we discarded 1560 tags, which were not present in $C$, from the initial 8128 ClarifAI tags.}. Each tag $t \in C$ is assigned a mean concreteness rating $c(t) \in [1,5]$, aggregated across at least 22 raters, where $1$ indicates highly abstract and $5$ indicates highly concrete. Examples of tags and their mean concreteness are: spirituality (1.07),  anniversary (3.15), kiss (4.48), actress (4.54), skin (4.79), people (4.82), machine gun (5.0).

To separate tags into concrete and abstract sets, we apply a threshold $4.75$. The threshold was chosen to include the maximum number of objects and the fewest other types of tags in the concrete set\footnote{While this separation relies on human labels and has some misalignments for tags with concreteness ranking close to 4.75 (e.g. tag 'football' is considered abstract, while 'soccer' is concrete), we observed only a few of such cases and consider them negligible due to the overall large tag number.}.
Thus, we define the abstract $A=\{t_j \in T | c(t_j) < 4.75\}$ and concrete $B=\{t_j \in T | c(t_j) \geq 4.75\}$ concept sets. For completeness, in the subsequent experiments, we also include the combined set $T = A \cup B$, which includes both abstract and concrete tags.

\section{Tag representation for classifier}
\label{sec: Tag representation for classifier}

The abstract $A$, concrete $B$, and combined $T$ tag sets differ in size ($|A| \gg |B|$), and all sets contain tags with highly varying detection frequencies across images in privacy datasets. However, a larger dictionary size or a greater number of tags used to describe an image can provide significantly more details about the content and, consequently, give an unfair advantage to a tag type. To ensure a fair comparison across concreteness types, we apply a two-step feature selection process for each set $S \in \{A, B, T\}$, namely discriminative selection, which restricts the dictionary size, and tag sparsity control, which ensures an equal number of tags for each image.

\begin{figure}[t]
    \begin{center}
    \includegraphics[width=0.47\textwidth]{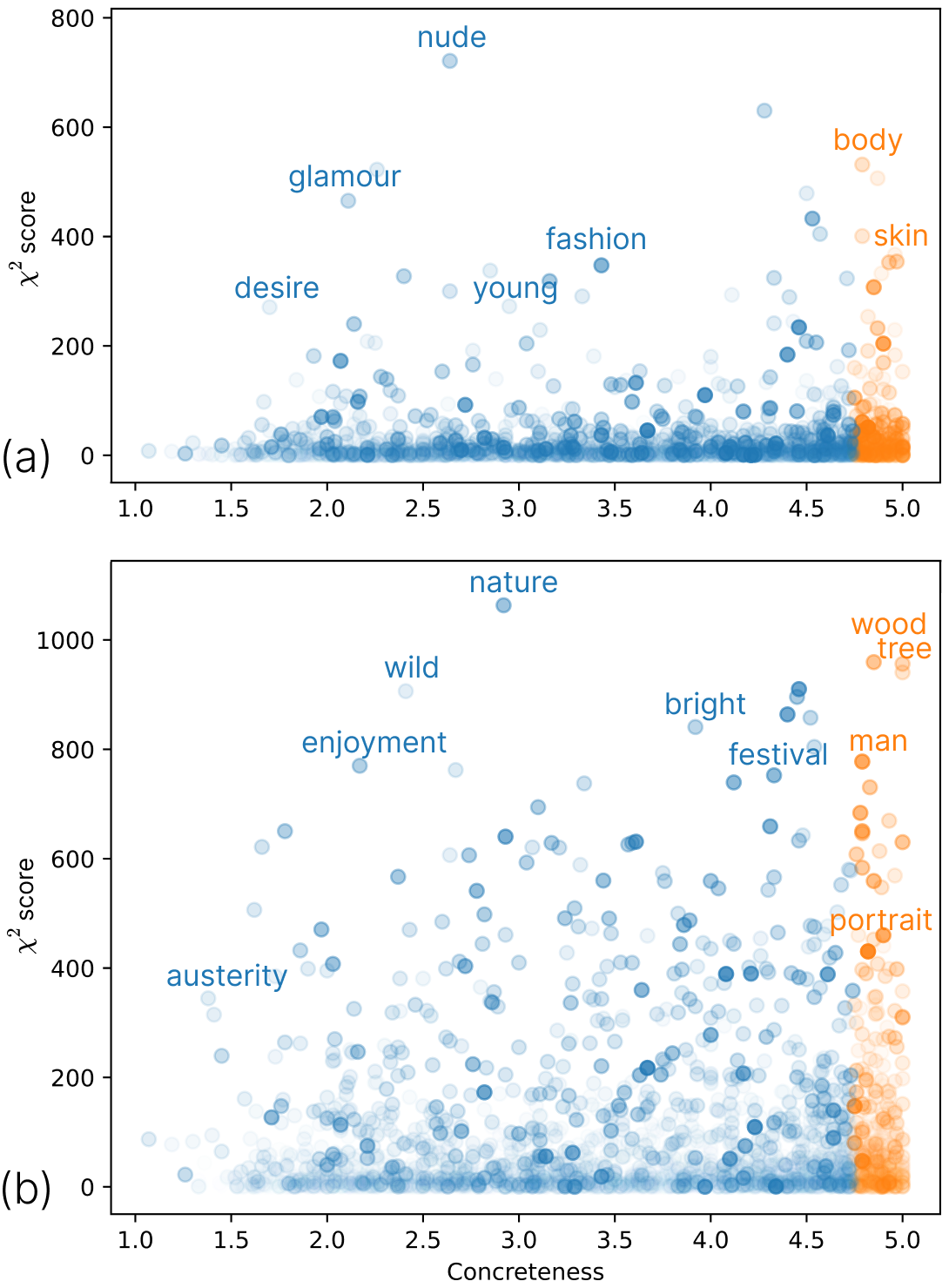}
    \end{center}
    \caption{Tag concreteness and their $\chi^2$ score with image privacy labels in the PrivacyAlert (a) and VISPR (b) datasets. The transparency of the dots reflects the tag frequency with which the tag appears in the dataset (the more transparent, the less frequent). We visualised the tag names of a few sample points. Both \textcolor{abstract_range_color}{abstract} and  \textcolor{concrete_range_color}{concrete} tag sets have tags with high $\chi^2$ scores, signalling that tags of different abstractness are important for image privacy.}
    \label{GT_concrete_1-5_dependancy_from_chi2_score_color-freq_size-std}
\end{figure}

\noindent{\textbf{Discriminative selection}. To equalise the dictionary sizes across all tag types, we calculate the $\chi^2$ score between the probabilities of each tag $t_j \in S$ appearing in the images from a dataset and their corresponding binary privacy labels. A higher $\chi^2$ score indicates a stronger correlation between the tag probability and the privacy value, suggesting that the tag is important for image privacy. We keep the top $M$ tags:
\begin{equation}
    S^\prime = top_M(\chi^2(S)) =  \{t^\prime_{1} , ..., t^\prime_{M}\}, 
\end{equation}
with $|S^\prime| = M = 1000$. Dictionary sizes around $1000$ are commonly used for image privacy~\cite{Zhao_2022_privacyalert, Zhao_2023_deep_gated, liu2023modality_coupling}. The resulting feature vector $p(x_i, S^\prime)$ for each image $x_i$ has dimension $M$. Fig.~\ref{GT_concrete_1-5_dependancy_from_chi2_score_color-freq_size-std} shows the distribution of tag concreteness and $\chi^2$ scores. It is notable that both concrete and abstract tag sets contain tags with high $\chi^2$ scores, indicating their potential importance for image privacy classification on the corresponding dataset.

\noindent{\textbf{Tag sparsity control}. To control the number of tags $k$ describing an image $x_i$, for each image we retain only the top-$k$ tags with the highest probabilities $p(x_i,t_j)$ among tags in $S^\prime$, setting all remaining tag probabilities to zero: 
\begin{equation}
p_k(x_i, t_j) = \Big\{ \begin{array}{ll}
  p(x_i, t_j), & \mbox{if } p(x_i, t_j) \in top_k\{p(x_i, S^\prime)\} \\
  0, & \mbox{otherwise}.
\end{array}
\end{equation}}
As a result for each tag type (set $S \in \{A, B, T\}$) image $x_i$ is described with a feature vector 
\begin{equation}
    p_k(x_i, S) = (p_k(x_i, t^{\prime}_1), ..., p_k(x_i, t^\prime_{M})). 
\end{equation}}
Therefore, for each image $x_i$ and number of tags $k$, we get three representations: through abstract $p_k(x_i, A)$, concrete $p_k(x_i, B)$, and combined $p_k(x_i, T)$ tags. Further, we investigate how the tag type and number affect the performance of classifiers trained on the corresponding representations. The selection of tag representation, described in this section, ensures that across all tag sets, the dictionary size and the number of active tags per image are matched, so that the only factor influencing performance is the abstractness of the tags.

\section{Experiments}

In this section, we train simple classifiers on abstract, concrete and combined tag representations while varying the number of tags per image, and analyse how these factors affect classification performance. Further, we also examine the co-occurrence patterns of abstract and concrete tags to provide insight into the observed performance trends.

For each tag type $S \in \{A, B, T\}$, the representation of image $x_i$ is $p_k(x_i, S)$ (Sec.~\ref{sec: Tag representation for classifier}). We explore different numbers of tags with which images are described, $k \in [1, 25]$. This allows us to inspect if the behaviour of the models on tags of various concreteness changes with the increase in details about an image that these tags provide.
If the image can not be described with as many as $k$ tags in at least one of the tag types, the image is disregarded from the experiments on $k$ tags. For $k=25$, only 5\% of images were disregarded from the PrivacyAlert and DIPA2 datasets (equally proportionally for train, val and test splits). However, for VISPR with $k=25$, 18\% of images in the dataset were discarded, and for $k=18$, around 5\%.

\begin{figure}[t!]
    \begin{center}
    \includegraphics[width=0.475\textwidth]{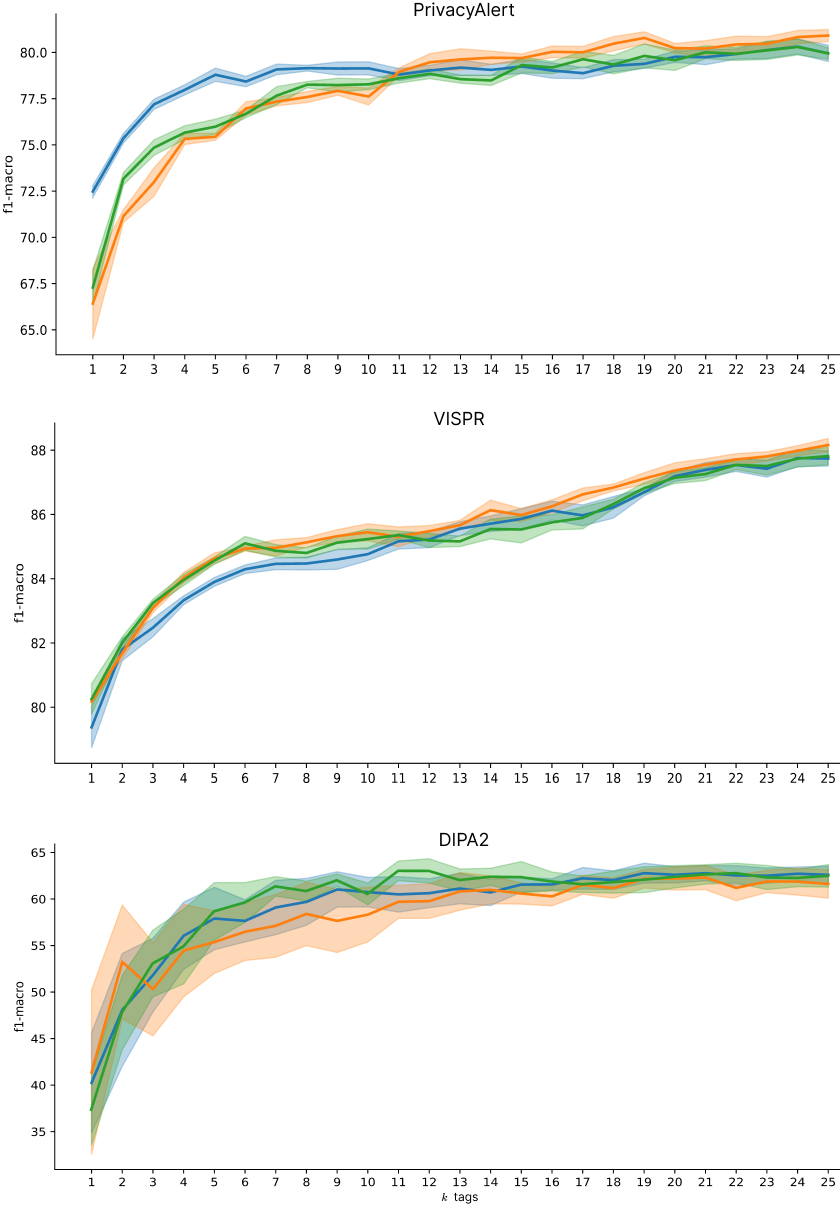}
    \end{center}
    \caption{The F1-macro (mean and std across 10 seeds) for MLP tag-based classifiers taking as input tags of different types: \textcolor{abstract_range_color}{abstract},  \textcolor{concrete_range_color}{concrete} and \textcolor{all_range_color}{combined}, with a varying number of tags used to describe an image. The performance is reported on three datasets: PrivacyAlert with subjective annotation, VISPR with object-guided annotation, and DIPA2 with subjective object-guided annotation. The presented F1-macro ranges vary in scale for each dataset.}
    \label{F1_macro_of_MLP_trained_on_n_tags_of_concreteness}
\end{figure}
\subsection{Models trained on different types of tags}
\label{sec: Model trained on different types of tags}
We first train a model taking as input tag-based feature representation of images $p_k(x_i, S)$ for different types $S$ and numbers of tags $k$.
We chose a simple two-layered MLP with two hidden layers of sizes 128 and 32 and a ReLU activation, as the model architecture for the analysis, similar to previous research~\cite{8features, xompero2025learningprivacyvisualentities}. 

For each dataset, hyperparameters were selected for $p_{14}(x_i, T)$ inputs, representing 14 tags describing each image with a combination of concrete and abstract tags, selected on 30 runs of Optuna\footnote{https://optuna.org}. Further, the same hyperparameters were used to train the models on this dataset for all numbers $k$ and types $S$ of tags. The performance is recorded across 10 seeds for each model.

Fig.~\ref{F1_macro_of_MLP_trained_on_n_tags_of_concreteness} shows the F1-macro of MLPs on concrete, abstract and combined tag image descriptions with varying levels of detail. 
For the subjective PrivacyAlert dataset, when the number of tags describing each image is small ($k \leq 10$), the model that takes as input abstract tags performs significantly better than the model on concrete tags by 2.80~p.p.~on average, and with 4.09~p.p.~average difference for $k \leq 5$. Additionally, the classifier taking as input 5 abstract tags performs as well as a classifier on 11 concrete tags. This behaviour on PrivacyAlert likely arises because abstract tags represent more contextually complex content, which helps better differentiate privacy. However, when the number of tags per image exceeds $10$, the models on all tag types converge to a similar performance. 

At the same time, for the object-related VISPR dataset, the performance of models on all three tag types is similar from the beginning, with a model on abstract tags on VISPR slightly underperforming by 0.62 to 0.75~p.p.~on average on 3 to 6 tags. Additionally, even on $k=1$ tag, the performance of the models on VISPR is only 8.37~p.p.~lower than their performance on 25 tags, while in PrivacyAlert, this gap is 14.50~p.p.~for concrete tags, which can be explained by the 'easier' nature of the VISPR classification task. 

On DIPA2 classifiers on abstract, concrete, and combined tag types perform more similarly than in PrivacyAlert. Additionally, on DIPA2, which combines object annotation with subjective perception of privacy of these objects, the best performance for the small $k$, especially for $k \leq 10$, is achieved by tags that combine concrete and abstract. This implies a relation between the subjectivity of labels and abstract concepts. The performance of models on DIPA2 is also characterised by a large standard deviation, attributable to performance instability arising from the limited number of images in the DIPA2 dataset.

Across all three datasets, performance increases with the number of tags, and in some cases, adding five tags to the description can yield up to 5~p.p.~improvements, highlighting the need for detailed, lengthy image descriptions when developing classifiers and explaining privacy decisions. Additionally, when the number of tags is sufficiently large ($k \geq 13$ across all datasets), the performance of models across all tag types becomes very similar, suggesting that they all convey related information about images.

\subsection{Analysis of tags co-occurrence}
We observed that when using a large number of tags, the model's performance was very similar for both concrete and abstract tags. To examine whether this effect arises from the direct co-occurrence of abstract and concrete tags, we compute the Jaccard index between every pair of abstract and concrete tags within each dataset. Specifically, we use feature vectors constructed for abstract  $p_{25}(x_i, A)$ and concrete $p_{25}(x_i, B)$ tag types with $k=25$ tags retained for each image. For each pair of tags $t_a \in A^\prime$ and $t_b \in B^\prime$, where $|A^\prime| = |B^\prime| = 1000$, the Jaccard index is defined as 
\begin{equation}
J(t_a, t_b) = \frac{\displaystyle\sum_{[x_i \in D_{tr}]}\hspace{-4pt}(\textbf{I}(x_i, t_a)\textbf{I}(x_i, t_b))}{\displaystyle\sum_{[x_i \in D_{tr}]}\hspace{-7pt} (\textbf{I}(x_i, t_a) + \textbf{I}(x_i, t_b) - \textbf{I}(x_i, t_a)\textbf{I}(x_i, t_b))}\text{,}
\end{equation}
where $\textbf{I}(x_i, t) = \textbf{I}_{[p_{25}(x_i, t) > 0]}$ indicates if $t$ is detected in the image $x_i$ from the train subset of the dataset $D_{tr}$.

\begin{table}[t]
\caption{Abstract - concrete tag pairs that have Jaccard index $> 0.5$ in image representation for each dataset.}
\vspace{7pt}
\begin{tabular}{llp{4.8cm}}
\toprule
\textbf{Dataset} & \textbf{\# pairs} & \textbf{Examples} \\
\midrule
PrivacyAlert & 120 & entrepreneur - cashier, automotive - mustang, smoker - ashtray, sexy - lingerie\\
VISPR &  99 & songbird - sparrow, pollination - bumblebee, adult - man, architecture - house\\
DIPA2 & 361 & manicure - fingernail, craftsmanship - sewing machine, herdsman - golden retriever, glacier - ski resort\\
\bottomrule
\multicolumn{3}{l}{\parbox{0.9\linewidth}{\scriptsize{Key -- \# pairs: number of abstract-concrete pairs}}}
\end{tabular}
\label{jaccard_index_for_concrete_and_abstract}
\end{table}

Table~\ref{jaccard_index_for_concrete_and_abstract} reports the number of abstract-concrete tag pairs that have a Jaccard index greater than $0.5$, indicating strong co-occurrence, along with the examples. For VISPR and PrivacyAlert, only around 100 concrete tags out of 1000 have a highly co-occurring abstract tag (out of 1000), suggesting that direct co-occurrence between concrete and abstract tags is uncommon. For DIPA2, the number of highly co-occurring pairs is higher, 352. However, due to the small size of this dataset, 54 of these pairs involve concepts that occur only once in the entire train set. In summary, these findings indicate that while abstract and concrete tags rarely exhibit strong direct co-occurrence, when an image is described with many tags, they begin to convey similar information about image privacy.

\section{Conclusion}

We analysed the importance of extracting diverse content tags for image privacy classification.
While a previous study suggested that abstract tags better represent images for privacy~\cite{galfre2019exploring_abstract_concepts}, we found that
as the number of tags per image description increases beyond 13, the performance gap between models trained on different tag types diminishes, and they begin to perform similarly. When only a small number of tags are available, the optimal choice of tag type becomes critical and is closely related to the subjectivity of the task: for object-guided annotation, concrete tags perform best, although abstract tags remain competitive; for subjective annotation, abstract tags provide richer information and outperform concrete ones.

Considering that tag-based or interpretable classification often explains decisions with a small number of tags, future research should incorporate not only object-oriented concepts but also abstract concepts, particularly for subjective tasks. However, since the detection of abstract tags may be more difficult~\cite{hessel2018quantifying_the_visual_concreteness}, with a large tag budget, concrete, object-related tags can replace abstract tags without a significant impact on the performance. Furthermore, classifier performance can benefit significantly from even modest increases in the detailedness of tag descriptions, especially for tasks where labelling is not strictly object-guided.

\bibliographystyle{IEEEbib}
\bibliography{main}

\end{document}